\documentclass{bmvc2k}

\usepackage{booktabs}
\usepackage{threeparttable}
\usepackage{multirow}
\usepackage{array}
\usepackage{wrapfig}

\title{Efficient Traffic-Sign Recognition with Scale-aware CNN}

\addauthor{Yuchen Yang}{linyeglasses@hotmail.com}{1}
\addauthor{Shuo Liu}{luke.liu@alumni.ubc.ca}{2}
\addauthor{Wei Ma*}{mawei@bjut.edu.cn}{1}
\addauthor{Qiuyuan Wang}{wangqiuyuan@pku.edu.cn}{3}
\addauthor{Zheng Liu}{zheng.liu@ubc.ca}{2}
\addinstitution{
 Beijing University of Technology\\
 100 Pingleyuan,\\
Chaoyang District, Beijing, China
}
\addinstitution{
 University of British Columbia, Okanagan\\
 3333 University Way,\\
Kelowna, BC, Canada
}
\addinstitution{
 Peking University\\
 5 Yiheyuan Road,\\
 Haidian District, Beijing, China
}

\runninghead{BMVC 2017}{Efficient Traffic-Sign Recognition with Scale-aware CNN}

\def\etal{\emph{et al}\bmvaOneDot}

\begin{document}

\maketitle

\begin{abstract}

The paper presents a Traffic Sign Recognition (TSR) system, which can fast and accurately recognize traffic signs of different sizes in images. The system consists of two well-designed Convolutional Neural Networks (CNNs), one for region proposals of traffic signs and one for classification of each region. In the proposal CNN, a Fully Convolutional Network (FCN) with a dual multi-scale architecture is proposed to achieve scale invariant detection. In training the proposal network, a modified "Online Hard Example Mining" (OHEM) scheme is adopted to suppress false positives. The classification network fuses multi-scale features as representation and adopts an "Inception" module for efficiency. We evaluate the proposed TSR system and its components with extensive experiments. Our method obtains $99.88\%$ precision and $96.61\%$ recall on the Swedish Traffic Signs Dataset (STSD), higher than state-of-the-art methods. Besides, our system is faster and more lightweight than state-of-the-art deep learning networks for traffic sign recognition. 

\end{abstract}

\section{Introduction}

Traffic sign recognition (TSR) is key to Advanced Driver Assistance Systems (ADAS) and Intelligent Transport Systems (ITS). Typically, given a road scene image, TSR automatically localizes and recognizes traffic signs in it, thereby reminding human drivers or helping ADAS make decisions. Many TSR methods have been proposed in recent years \cite{Ciresan1,Ciresan2,practical,fcn-edgebox,yang2016towards,wang2017traffic}, and validated on some public traffic sign recognition benchmarks \cite{GTSDB,GTSRB,STSD}.


However, there are still many challenging problems for real-world applications. First, road scenes are extremely complicated, because of varying illumination, color deterioration of traffic signs, and the existence of decorations looking similar to traffic signs. In this situation, it is hard for TSR to obtain a high recall rate while producing few false positives during detection. Second, unlike lab environments, automobiles have limited memory and computing capacity. Thus, the computational complexity of TSR should be low and its required memory resources should be small. Third, the scales of traffic signs captured by moving vehicles vary in a large range. Traffic signs, especially small ones, are easily missed by current TSR systems.

\begin{figure*}[htp]
\centering
\includegraphics[width=0.8\textwidth]{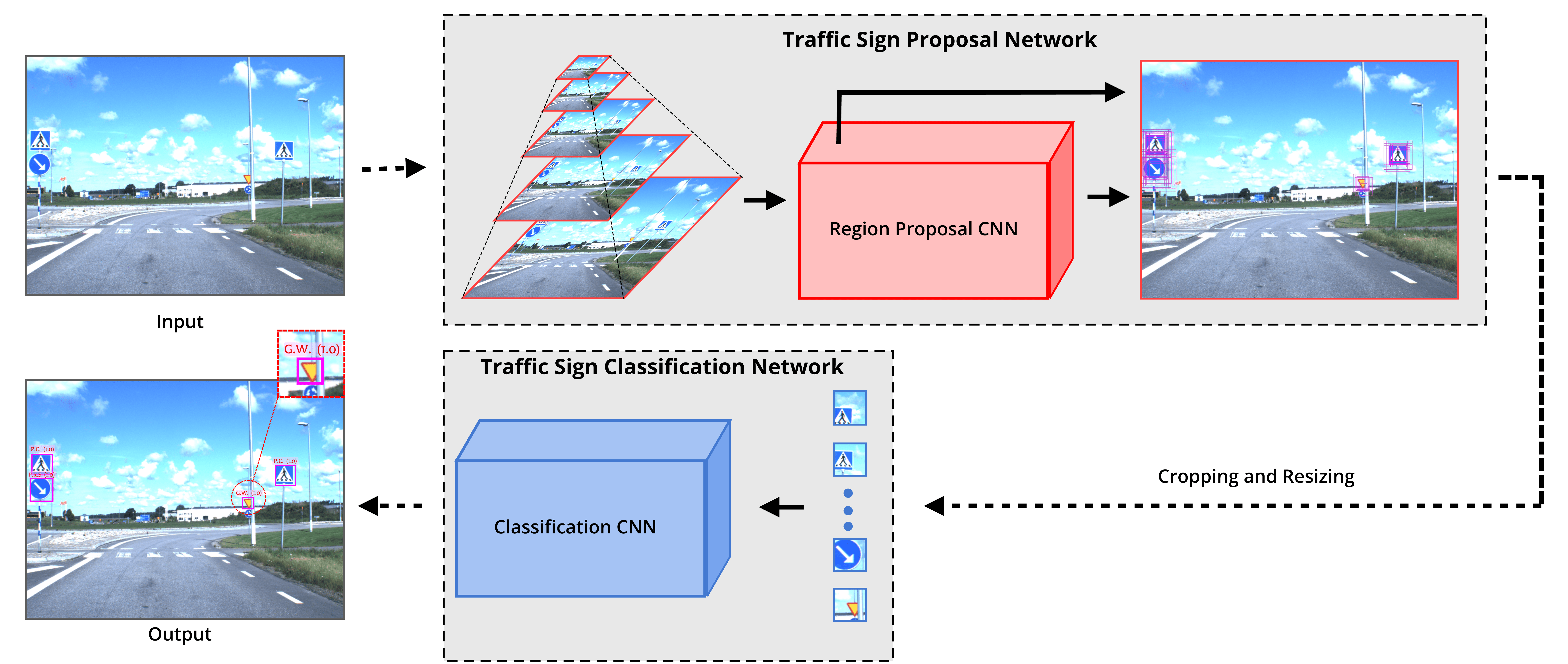}
\caption{Overview of the proposed scale-aware TSR system.}
\label{overview}
\end{figure*}

In this paper, we propose a scale-aware Traffic Sign Recognition framework (scale-aware TSR for short), which deals the above problems well. The overview of the proposed framework is illustrated in Figure \ref{overview}. It consists of two Convolutional Neural Networks (CNNs), a proposal (class-agnostic detection) network to find possible regions of traffic signs and a classification network for classifying each region proposal. The proposal network adopts a dual multi-scale structure. Given an image, a rough image pyramid is constructed beforehand and fed into the network. Inside of the network, there are two output branches, one from a lower layer for small-scale objects and the other from the last layer for large-scale objects. The results of the dual multi-scale proposal network are aggregated and sent to the classification network. The classification network is designed to be lightweight and capable of fusing multi-scale features.


Our major contributions are as follows: (1) A traffic sign recognition framework named scale-aware TSR is proposed. It beats state-of-the-art methods and achieves $99.88\%$ precision and $96.61\%$ average recall rate on Swedish Traffic Signs Dataset (STSD) \cite{STSD}; (2) Within the framework, we elaborately design a Fully Convolution Network (FCN) with an efficient dual multi-scale structure for region proposals. The network is experimentally verified to be scale-aware, thereby capable of localizing targets of wide-range scales; (3) Moreover, a lightweight classification sub-network with multi-scale feature fusion structure and "Inception" module \cite{inception} is presented and validated.

\section{Related Work}
\label{sec:relate}


In traditional paradigm, TSR is usually carried out as two separate sub-tasks: class-agnostic detection and class-specific classification. They are validated on separate datasets, e.g. the German Traffic Sign Detection Benchmark (GTSDB) \cite{GTSDB} for detection and the German Traffic Sign Recognition Benchmark (GTSRB) \cite{GTSRB} for classification. 


In the task of traffic sign detection, most state-of-the-art methods \cite{slidewindow,shapeAcolor,GTSDB} adopted a general pipeline where a "sliding window" is firstly employed, followed by a hand-crafted feature extractor such as HOG \cite{HOG} or Hough \cite{Hough}, and a classifier (SVM \cite{SVM} or Random Forests \cite{RandomForests}). As the sliding window scheme involves exhaustive search, these methods are too time-consuming to be deployed in real-world scenarios. To tackle this issue, several Regions of Interest (ROIs) based methods \cite{ROIextration,salti2013traffic,yang2016towards} were proposed to replace the exhaustive search and gain acceleration. In particular, \cite{yang2016towards} proposed a color probability model to estimate a probability map from an input image. Based on the probability map, a Maximally Stable Extremal Regions (MSERs) detector is applied to extract traffic sign ROI proposals. This method costs only 67$ms$ per image in GTSDB test dataset. These hand-crafted detectors might not have a promising recall rate when they are applied in new scenes. Most recently, Aghdam \etal \cite{practical} combined sliding window, feature extractors and classifiers into an end-to-end CNN architecture, which gains improvement on both accuracy and speed. Note that our traffic sign proposal network is partly motivated by this approach.


In the traffic sign classification task, most current methods are based on the assumption that all possible traffic signs have already been detected successfully and their locations are accurate. Before CNNs started to lead state-of-the-art performance, various traditional hand-engineered features along with shallow classifiers were exploited for traffic sign classification, such as pixel-level features + SVM ~\cite{Maldonado-Bascon} and HOG + Random Forests~\cite{Zaklouta}. Recently, \cite{GTSRB} and \cite{GTSDB} showed that CNN-based methods achieved better-than-human classification rate. In \cite{Ciresan1,Ciresan2}, a committee of CNNs was proposed, where a CNN and a Multi-Layer Perceptron (MLP) are combined together and trained on raw pixel intensities. They obtained a high recognition rate up to $99.15\%$ on GTSRB. Furthermore, \cite{jin2014traffic} proposed a hinge loss stochastic gradient descent method to train CNNs and achieved $99.65\%$ recognition rate on GTSRB. To balance the accuracy and computational efficiency, a small scale CNN was proposed in \cite{fcn-edgebox}. It adopted bootstrapping training to enhance its discriminative ability. \cite{haloi2015traffic} improved the performance on GTSRB to $99.81\%$ with the combination of the insight from the inception module \cite{inception} and the spatial transformer networks \cite{jaderberg2015spatial}. However, these approaches are not lightweight enough for hardware equipment mounted on vehicles. In addition, as mentioned earlier, these methods are tested on already perfectly detected traffic signs. We consider traffic sign class-agnostic detection and classification together for practical applications.

We also researched methods dealing with detection and classification simultaneously, most of which target at generic objects. The most popular framework among them is R-CNN \cite{rcnn}. It uses a pre-trained CNN to extract features from box proposals generated by selective search \cite{selective_search}, and then class-specific linear SVMs for classification. The significant advantage of this work is the replacement of hand-engineered features with CNN extracted ones. Meanwhile, some variants of R-CNN were proposed to decrease the running time of R-CNN \cite{he2014spatial,girshick2015fast,fasterrcnn}. It is worth noting that Faster R-CNN \cite{fasterrcnn} amended the multi-stage task in R-CNN into an end-to-end detection system which consists of a Region proposal network (RPN) as well as a CNN for classification and bounding box regression. Most recently, some TSR works gained insights from generic object detection and developed end-to-end systems. For example, motivated by OverFeat \cite{sermanet2013overfeat}, Zhu \etal \cite{zhu2016traffic} proposed an FCN that can detect and classify traffic signs simultaneously. \cite{fcn-edgebox} used FCN \cite{FCN} and EdgeBoxes \cite{edge-boxes} to generate region proposals of traffic signs in a coarse-to-fine manner, and then a small size CNN to classify these proposals. This method was evaluated on STSD which is designed for testing the performance on entire TSR systems. We treat this method \cite{fcn-edgebox} as our baseline and also evaluate our method on STSD.

\section{Methodology}

In the following, we describe the two parts in the proposed TSR system (shown in Figure \ref{overview}), region proposal network and classification network, respectively.

\begin{figure}[htp]
\centering
\includegraphics[width=0.8\textwidth]{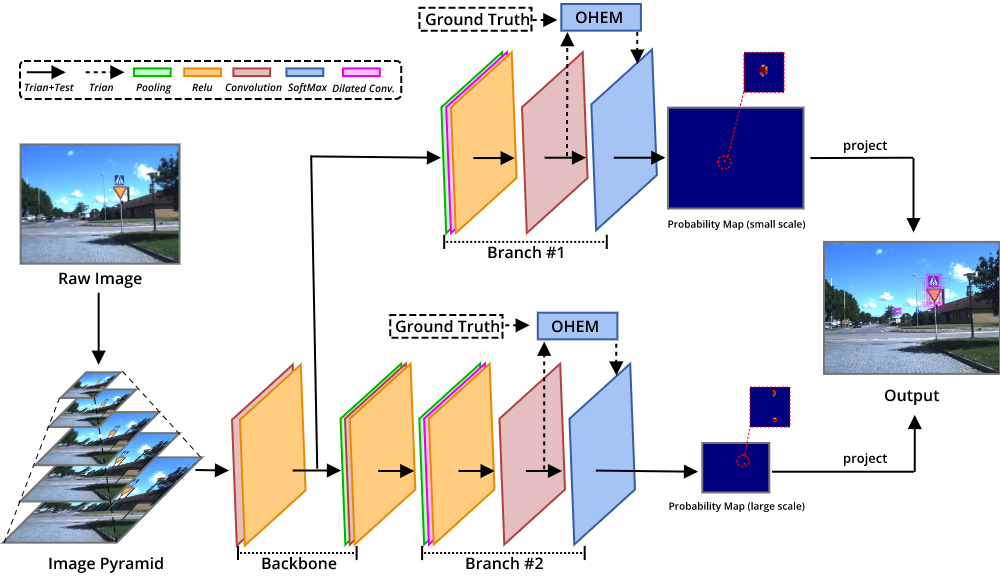}
\caption{Flowchart and architecture of the proposed DMS-net}
\label{proposal}
\end{figure}
\subsection{Proposal Network}



A dual multi-scale network (DMS-net) is designed for region proposals of traffic signs. Figure \ref{proposal} illustrates the flowchart and architecture of the DMS-net. Initially, a pyramid with multi-scale images is constructed based on a raw image. The multi-scale pyramid is forwarded through a hierarchical convolutional network layer by layer. For each layer of the image pyramid, the network produces multi-scale probability maps, in which a value represents the probability of its corresponding region in the input image being a part of a traffic sign. The dual multi-scale probability maps are aggregated to determine traffic sign proposals in the form of bounding boxes in the raw image.


\textbf{Dual Multi-scale Network:} Localizing objects in different scales is essential for an outstanding object detector in real-world applications. There are two main strategies to achieve this goal. The first is to train a detector with objects of various scales so that it can localize objects of different sizes in testing images \cite{slidingwindow_MS}. An alternative approach is to construct a hierarchical structure inside of the detector. Through the structure, the detector draws features of multiple scales to help make decisions. Several state-of-the-art detection methods adopted this strategy, such as MSCNN \cite{mscnn}. In the proposed DMS-net, we combine these two strategies together in the framework of CNN in an efficient way.


As seen in Figure \ref{proposal}, the DMS-net consists of two multi-scale structures, a rough image pyramid (5 scales in our experiments) and a hierarchical network with two output branches. In this way, our DMS-net can detect traffic signs at 10 different scales ($5 \times 2$) in total with high computation efficiency.


\textbf{Architecture:} The DMS-net adopts an FCN structure which can take an image of arbitrary size as input. This makes our method possible to deal with image pyramid. The network primarily includes a "backbone" stream and two "branches". To be specific, there are two convolution layers in the "backbone". The first one is followed by a Relu \cite{RELU} activation function, and the second one is similar except being preceded by a max pooling layer. The two branches have the same structure with a minor difference in configuration. The structure includes a max pooling layer, a dilated convolution layer \cite{dilation} with the Relu, as well as a $1\times1$ convolution layer and softmax layer for probability maps. Note that the first branch is out from the first layer of the backbone and the second comes from the end of the backbone. The detailed configurations of our DMS-net are given in Table \ref{networks_details}.

\begin{table}
\centering

\resizebox{\textwidth}{!}{
\begin{tabular}{|c|c|c|c|c|c|c|c|c|c|c|c|c|c|}
\hline
\multicolumn{14}{|c|}{Proposal Network (DMS-net)}                                                                                                                                                                                                                                                                                                                                                                                                                                                                                                                                                                                                                                                                                                                                                                                                                                                                                                                                                                 \\ \hline
Module Name & \multicolumn{4}{c|}{Backbone}                                                                                                                                                                                                                                                                                                                  &                                                                                    & \multicolumn{4}{c|}{Branch \#1}                                                                                                                                                                                                                                                                       & \multicolumn{4}{c|}{Branch \#2}                                                                                                                                                                                                             \\ \cline{1-5} \cline{7-14} 
Layer       & \multicolumn{2}{c|}{Conv1}                                                                                                                                            & \multicolumn{2}{c|}{Conv2}                                                                                                                                             &                                                                                    & \multicolumn{2}{c|}{Conv1\_1}                                                                                                                     & \multicolumn{2}{c|}{Conv1\_2}                                                                                                                     & \multicolumn{2}{c|}{Conv2\_1}                                                                                                                              & \multicolumn{2}{c|}{Conv2\_2}                                                  \\ \cline{1-5} \cline{7-14} 
Details     & \multicolumn{2}{c|}{\begin{tabular}[c]{@{}c@{}}C(60, 9, 9)\\ st.1, Relu\end{tabular}}                                                                                 & \multicolumn{2}{c|}{\begin{tabular}[c]{@{}c@{}}P(2, 2)\\ C(120, 5, 5)\\ st.1, Relu\end{tabular}}                                                                       &                                                                                    & \multicolumn{2}{c|}{\begin{tabular}[c]{@{}c@{}}P(6, 2)\\ C(300, 2, 2)\\ dilation 3\\ st.1, Relu\end{tabular}}                                     & \multicolumn{2}{c|}{\begin{tabular}[c]{@{}c@{}}C(2, 1, 1)\\ st.1\end{tabular}}                                                                    & \multicolumn{2}{c|}{\begin{tabular}[c]{@{}c@{}}P(4, 2)\\ C(300, 3, 3)\\ dilation 2\\ st.1, Relu\end{tabular}}                                              & \multicolumn{2}{c|}{\begin{tabular}[c]{@{}c@{}}C(2, 1, 1)\\ st.1\end{tabular}} \\ \hline
\multicolumn{14}{|c|}{Classification Network (fusion-net)}                                                                                                                                                                                                                                                                                                                                                                                                                                                                                                                                                                                                                                                                                                                                                                                                                                                                                                                                                           \\ \hline
Module Name & \multicolumn{2}{c|}{Basic Layers}                                                                                                                                     & \multicolumn{3}{c|}{Multi-Scale Features Fusion Module}                                                                                                                                                                                                     & \multicolumn{6}{c|}{Inception Module}                                                                                                                                                                                                                                                                                                                                                                                                                              & \multicolumn{2}{c|}{Classifier Layers}                                         \\ \hline
Layer       & Conv1                                                                             & Conv2                                                                             & Conv3\_1                                                                          & Conv3\_2                                                                           & Conv3\_3                                                                           & \#1*1                                                                   & \#3*3 Reduce                                                            & \#3*3                                                                   & \#5*5 reduce                                                            & \#5*5                                                                   & pool project                                                                     & Avg Pool                                  & FC                                 \\ \hline
Details    & \begin{tabular}[c]{@{}c@{}}C(100,5,5)\\ st. 1, pad 2\\ Relu\\ P(2,2)\end{tabular} & \begin{tabular}[c]{@{}c@{}}C(150,3,3)\\ st. 1, pad 1\\ Relu\\ P(2,2)\end{tabular} & \begin{tabular}[c]{@{}c@{}}C(250,3,3)\\ st. 1, pad 1\\ Relu\\ P(2,2)\end{tabular} & \begin{tabular}[c]{@{}c@{}}C(250,3,3)\\ st. 1, pad 1\\ Relu\\ P(2, 2)\end{tabular} & \begin{tabular}[c]{@{}c@{}}C(250,3,3)\\ st. 1, pad 1\\ Relu\\ P(2, 2)\end{tabular} & \begin{tabular}[c]{@{}c@{}}C(64,1,1)\\ st. 1, pad 1\\ Relu\end{tabular} & \begin{tabular}[c]{@{}c@{}}C(32,1,1)\\ st. 1, pad 1\\ Relu\end{tabular} & \begin{tabular}[c]{@{}c@{}}C(32,3,3)\\ st. 1, pad 1\\ Relu\end{tabular} & \begin{tabular}[c]{@{}c@{}}C(16,1,1)\\ st. 1, pad 1\\ Relu\end{tabular} & \begin{tabular}[c]{@{}c@{}}C(32,5,5)\\ st. 1, pad 1\\ Relu\end{tabular} & \begin{tabular}[c]{@{}c@{}}P(3,1)\\ C(96,1,1)\\ st. 1, pad 1\\ Relu\end{tabular} & Ave\_P(5,1)                               & FC 11                              \\ \hline
\end{tabular}
}
\caption{The detailed settings of our DMS-net (top) and fusion-net(bottom). In the details cell, "C (number,size,size)" denotes the number of convolution filters and their local receptive size; "st." and "pad" indicate the convolution stride and spatial padding, respectively. "P" indicates average pooling with kernel size and stride.}
\label{networks_details}
\end{table}

\textbf{Dense Prediction:} Another reason to use the FCN structure is its ability to make dense predictions for per-pixel tasks like semantic segmentation \cite{FCN}. Unlike semantic segmentation \cite{dilation,FCN} where each output pixel is a classifier of its path-connected one in the input image, each value in our probability maps represents the probability of its corresponding receptive field in the input image as parts of traffic signs. In this way, we can generate region proposals with the FCN structure. This idea is mainly motivated by the works in \cite{practical,dilation}.

We take the first branch in Figure \ref{proposal} and its output as an example to explain the correspondences between probability maps and receptive fields in the input image. Given an image patch of $20 \times 20$ pixels, the first convolution layer with kernels of size $9 \times 9$ generates $12 \times 12$ feature maps. In these feature maps, each value has a $9 \times 9$ receptive field in the original image. Then, after the max pooling layer in the first branch with kernels of size $6 \times 6$ and stride of 2, the feature maps become $4 \times 4$ with a receptive field of size $14 \times 14$. Note that the first convolution layer in the branch is dilated convolution. The dilated convolution has been proved to support exponentially expanding receptive fields without losing resolution or coverage \cite{practical,dilation}. For more details about dilated convolution, please refer to \cite{dilation}. In this way, the dilated convolution layer with kernels of size $2 \times 2$ and dilation factor of 3 will produce only one value which has a receptive field of the whole image patch ($20 \times 20$). The last $1 \times 1$ convolution layer and softmax producing probability maps will not affect the receptive field. Therefore, for an arbitrary scale image, each value in the probability maps of the first branch corresponds to a region of size $20 \times 20$ pixels in the image.

Since there is an additional convolution layer and max pooling layer, the second branch has a larger receptive field. By projecting the multi-scale probability maps into the pyramid images and aggregating them, we obtain dense multi-scale bounding boxes. Note that in our DMS-net, we ignore the bounding boxes with irregular aspect ratios and only keep the square ones. This is because of the observation that most traffic signs stay within square regions.

\textbf{Training:} In order to train the DMS-net, we transform the original ground truth (bounding box style) into spatial probability maps by scanning each image with a square window. If the square window has an intersection-over-union (IoU) with traffic signs higher than 0.7 threshold, we assign $1$ (positive) to the objective probability map. If the IoU is lower than $0.3$, we assign $0$ (negative) to the probability map. Otherwise, we assign a special value, which will be ignored in loss calculation. The size of the scanning window depends on the receptive fields of the specific branches. For each image, we generate 10 ground truth maps with different scales in all. In order to provide a sufficient data to train our network, we adopt several data augmentation tricks including contrast adjustments, random blur, and lighting adjustments. Note that unlike the common patch-level training paradigm in pixel-to-pixel tasks \cite{patch1,patch2}, we input the whole images for training efficiently. Because the ratio of negative/positive in the ground truth maps is extremely high, the overall loss will be too small to be backpropagated if we sum over pixel-wise losses in the probability maps. To solve this issue, we adopt the modified Online Hard Example Mining (OHEM) \cite{OHEM} in the training stage to suppress false positives. To be specific, we calculate the pixel-wise loss on all the positions of probability maps. Then we sort the positions by their loss values and select top N positions ($N = 128$ in our experiments). Only the top N selected positions participate in the back-propagation process. This technique guarantees that our DMS-net converges fast and generates few false positives. We train the two branches in our network in a multi-stage way. Specifically, only one branch is trained while the other branch is frozen in each iteration. For training, we use same standard cross-entropy loss function for each branch with Stochastic Gradient Descent(SGD). The layers are initialized from a Gaussian distribution with zero mean and $0.01$ variance. Other hyper-parameters are momentum $0.9$; weight decay $0.0005$; batch size $1$; initial learning rate $0.001$, which is decreased by a factor of $10$ after $20k$ iterations. We trained the model for $50k$ iterations in our experiments.

\subsection{Classification Network}

After the region proposals are extracted, we use a classification network, called fusion-net, to differentiate traffic signs of different classes and background. The fusion-net comprises two basic components, a multi-scale feature fusion module and an "Inception" module \cite{inception}, followed by an average pooling layer and a fully connected layer. The input image size is $64 \times 64$. The details about our fusion-net are listed in Table \ref{networks_details}.


\textbf{Multi-Scale Features Fusion Module:} Fusing features from different layers in a CNN is proved to be effective in improving accuracy in many tasks \cite{FCN,Sermanet,hypernet,inside-outside}. Features from different layers have different receptive fields and semantic clues, thereby helping our classification network differentiate fine-grain classes, e.g different speed signs. Our multi-scale feature fusion module has three unified $3 \times 3$ convolution layers. The output of each layer is branched out and then concatenated together.


\textbf{Inception Module and Average Pooling:} The "Inception" module \cite{inception}  performs multi-size convolution at the same time. All the results are then concatenated. This allows the model to take advantage of multi-level feature extraction from the same input with less time consumption and parameters. For more details about the "Inception" module, please refer to Table \ref{networks_details} and \cite{inception}. After the "Inception" module, we perform a global average pooling operation as a bridge to connect later with a fully connected layer. This can dramatically help fusion-net reduce overall parameters and prevent over-fitting.

\textbf{Training:} The fusion-net is not trained with DMS-net in an end-to-end manner. Similar to the method in \cite{fcn-edgebox}, we adopt bootstrapping to mine hard negative for training our fusion-net. Instead of sampling training data from the region proposal results, we randomly crop square regions, which have IoU with the traffic signs higher than 0.6 on the raw training images, as positive training data. Regions with the IoU lower than 0.5 are treated as negative training data. After the convergence of the network training, we use the model to find the wrongly classified samples and add them to the training set to further optimize the network. Most of the hyper-parameters are the same with those in training the proposal network, except that the batch size is 128, the learning rate is decreased by a factor of 5 after 20k iterations and the training ends after 100k iterations.


\textbf{Post Processing:} In the testing stage, we feed the bounding boxes generated by the DMS-net to the fusion-net. Then we perform non-maximum suppression over all the region proposals according to their classification scores. Finally, we adopt the bounding boxes voting trick \cite{boxvote} to further boost the location accuracy.

\section{Experiments}
The proposed scale-aware TSR system is implemented in the framework of Caffe \cite{caffe}, and run on a machine with a 4-core CPU@2.7GHz, 16G RAM, and a NVIDIA K40 GPU. We evaluate our method on STSD \cite{STSD}, which consists of more than 2,000 annotated images. We follow the experimental configurations in \cite{fcn-edgebox} to split the training and testing data.


\subsection{Evaluation on Standard STSD Benchmark}
We compare our system with state-of-the-art methods \cite{STSD,fcn-edgebox,fasterrcnn,rcnn,adaboost_svm} based on the standard evaluation benchmark proposed in \cite{STSD}, where only traffic signs labeled as 'visible' and larger than $50 \times 50$ are considered. The performances of the methods are measured by precision and recall separately, where results having IoU with ground truth higher than 0.5 are treated as true positive.

\begin{table}
\begin{center}
\tiny
\begin{tabular}{lp{0.25cm}p{0.25cm}p{0.25cm}p{0.25cm}p{0.25cm}p{0.25cm}p{0.25cm}p{0.25cm}p{0.25cm}p{0.25cm}p{0.25cm}p{0.25cm}}

    \toprule
    \multirow{2}{*}{Sign name}&\multicolumn{2}{c}{ FDs\cite{STSD} }&\multicolumn{2}{c}{ Adaboost+SVR\cite{adaboost_svm}}&\multicolumn{2}{c}{ R-CNN\cite{rcnn}}&\multicolumn{2}{c}{ Faster R-CNN\cite{fasterrcnn} }&\multicolumn{2}{c}{ FCN+EdgeBoxes\cite{fcn-edgebox}}&\multicolumn{2}{c}{ scale-aware TSR (ours)}\cr
    \cmidrule(lr){2-3} \cmidrule(lr){4-5} \cmidrule(lr){6-7}\cmidrule(lr){8-9}\cmidrule(lr){10-11}\cmidrule(lr){12-13}
    &Prec.($\%$)&Rec.($\%$)&Prec.($\%$)&Rec.($\%$)&Prec.($\%$)&Rec.($\%$)&Prec.($\%$)&Rec.($\%$)&Prec.($\%$)&Rec.($\%$)&Prec.($\%$)&Rec.($\%$)\cr
    \midrule
    \tiny {PEDESTRAIN CROSSING}
    &96.03	&91.77	&98.52	&93.45	&87.9	&87.2	&94.21	&97.44	&{\bf100}	&95.20	&{\bf100}	&{\bf99.15}\cr
    \tiny {PASS RIGHT SIDE}
	&{\bf100}	&95.33	&{\bf100}	&97.53	&93.8	&93.8	&94.44	&96.23	&95.3	&93.8	&{\bf100}	&{\bf100}\cr
    \tiny {NO STOPPING NO STANDING}
	&97.14	&77.27	&99.20	&81.46	&66.8	&71.7	&85.71	&54.55	&{\bf100}	&75.0	&{\bf100}	&{\bf100}\cr
    \tiny {50 SIGN}
	&{\bf100}	&76.12	&{\bf100}	&80.56	&{\bf100}	&{\bf100}	&{\bf100}	&{\bf100}	&{\bf100}	&{\bf100}	&{\bf100}	&{\bf100}\cr
    \tiny {PRIORITY ROAD}
	&98.66	&47.76	&97.89	&79.68	&95.7	&97.8	&96.51	&98.81	&{\bf100}	&{\bf98.9}	&98.77	&95.24\cr
    \tiny {GIVE WAY}
	&59.26	&47.76	&71.50	&52.39	&79.4	&90		&{\bf100}	&85.71	&96.7	&{\bf96.7}	&{\bf100}	&96.43\cr
	\tiny {70 SIGN}
	&-		&-		&-		&-		&92.6	&86.2	&{\bf100}	&{\bf100}	&{\bf100}	&{\bf100}	&{\bf100}	&{\bf100}\cr
	\tiny {80 SIGN}
	&-		&-		&-		&-		&{\bf100}	&77.3	&{\bf100}	&{\bf100}	&94.4	&77.3	&{\bf100}	&95.24\cr
	\tiny {100 SIGN}
	&-		&-		&-		&-		&{\bf100}	&{\bf100}	&{\bf100}	&73.68	&90.5	&{\bf100}	&{\bf100}	&94.74\cr
    \tiny {NO PARKING}
	&-		&-		&-		&-		&96.3	&68.4	&{\bf100}	&76.47	&{\bf100}	&{\bf92.1}	&{\bf100}	&85.29\cr
    \cr
    Average (the first 6 classes)
	&91.84	&77.08	&94.52	&80.85	&90.08	&87.27	&95.15	&88.79	&98.67	&93.27	&{\bf99.79}	&{\bf98.47}\cr
    Average (all)
	&-		&-		&-		&-		&91.25	&87.24	&97.09	&88.29	&97.69	&92.90	&{\bf99.88}	&{\bf96.61}\cr
    
    \bottomrule
\end{tabular}
\end{center}
\caption{Accuracy comparison on STSD based on standard evaluation benchmark}
\label{accuracy_standard}
\end{table}

\begin{table}
\begin{center}
\tiny
\begin{tabular}{lp{0.5cm}<{\centering}rp{0.5cm}<{\centering}rp{0.5cm}<{\centering}rp{0.5cm}<{\centering}r}

    \toprule
    \multirow{2}{*}{Methods}&\multicolumn{2}{c}{ Region Proposal }&\multicolumn{2}{c}{ Classification }&\multicolumn{2}{c}{ Overall}&\multicolumn{2}{c}{ Accuracy}\cr
    \cmidrule(lr){2-3} \cmidrule(lr){4-5} \cmidrule(lr){6-7}\cmidrule(lr){8-9}
    &Time(s/img)	&Parameters(million)	&Time(s/img)	&Parameters(million)	&Time(s/img)	&Parameters(million) &Prec.($\%$)	&Rec.($\%$) \cr
    \midrule
    FCN+EdgeBoxes\cite{fcn-edgebox}
    &0.578	&14.7 	&0.052	&53.4 	& $\approx$0.63 &68.1  & 97.69 &92.90\cr
    Faster R-CNN \cite{fasterrcnn}
	&0.423	&17.1 	&\textbf{0.041}	&119.7	& 0.511  &136.8 &97.09 &88.29 \cr
    \hline
    scale-aware TSR
	&\textbf{0.226}	&\textbf{0.5}	&0.102	&\textbf{1.8}	& 0.413	&\textbf{2.4} &\textbf{99.88}	&\textbf{96.61} \cr
    scale-aware TSR (64 proposals)
	&\textbf{0.226}	&\textbf{0.5} 	&0.056s	&\textbf{1.8} 	& \textbf{0.356}	&\textbf{2.4} &\textbf{99.88}	&96.52 \cr
    \bottomrule
\end{tabular}
\end{center}
\caption{ Comparison of running time and model complexity. }
\label{efficency}
\end{table}


\textbf{ Comparison on Accuracy:}  Table \ref{accuracy_standard} gives precisions and recall rates of FDs \cite{STSD}, Adaboost+SVR \cite{adaboost_svm}, R-CNN \cite{rcnn}, Faster R-CNN \cite{fasterrcnn}, FCN+EdgeBoxes \cite{fcn-edgebox} and our scale-aware TSR. For Faster R-CNN, we use VGG16 as its backbone network and $960 \times 1280$ as input size. All the training and testing settings are the same with the original paper \cite{fasterrcnn}. For all the experiments except those specifically stated ones, we truncate 128 proposals generated by DMS-net. As shown in table \ref{accuracy_standard}, our method achieves an average precision of $99.79 \%$ and an average recall of $98.33\%$ on the first 6 classes, which are much better than those of any other methods. Most notably, the improvement on the recall rate, $5.06$ points higher than the best method \cite{fcn-edgebox}, is significant for traffic sign recognition. We argue that it mainly gains from the dual multi-scale design in the DMS-net, which will be verified in the next section. For all the classes, we also have the best performance ($99.88\%$ precision and $96.61\%$ recall rate). In comparison with the baseline method \cite{fcn-edgebox}, we gain $2.19\%$ and $3.71\%$ improvement on precision and recall rate, respectively. Moreover, we observe that the classification of all types of speed signs generated by our proposal network are $100\%$ correct, due to the multi-scale feature fusion structure in the fusion-net enabling fine-grained recognition. 


\textbf{Comparison on Efficiency:} Fast computation and low model complexity are essential for practical applications. We take the top 3 methods (Faster R-CNN \cite{fasterrcnn}, FCN+EdgeBoxes \cite{fcn-edgebox} and ours) in Table \ref{accuracy_standard} for model complexity and time cost comparison. Due to the lightweight consideration in system design, the number of parameters in our entire system is around $57\times$ lower than that in Faster R-CNN and around $28\times$ lower than that in \cite{fcn-edgebox}. Therefore, our system is better suited to automobiles with on-board hardware. In order to test the time costs, we run the above three methods with their default settings on our machine whose configuration was presented earlier. Since the overall code of FCN+EdgeBoxes \cite{fcn-edgebox} is not public, we reproduce their work and estimate their time costs on our machine. The time may have a minor difference with that presented in their paper due to different machines. We show the speeds of each stage and the overall systems in Table \ref{efficency}. Although our classification is slower than that of the others, our whole system is the fastest one ($0.413s$ in total running time). Furthermore, We also attempt to reduce the region proposals from 128 to 64. Our method gains $0.057s$ speedup, while it only drops $0.09\%$ recall rate.





\subsection{Ablation Study}
\begin{wrapfigure}{r}{5cm}

\centering
\includegraphics[width = 0.4\textwidth]{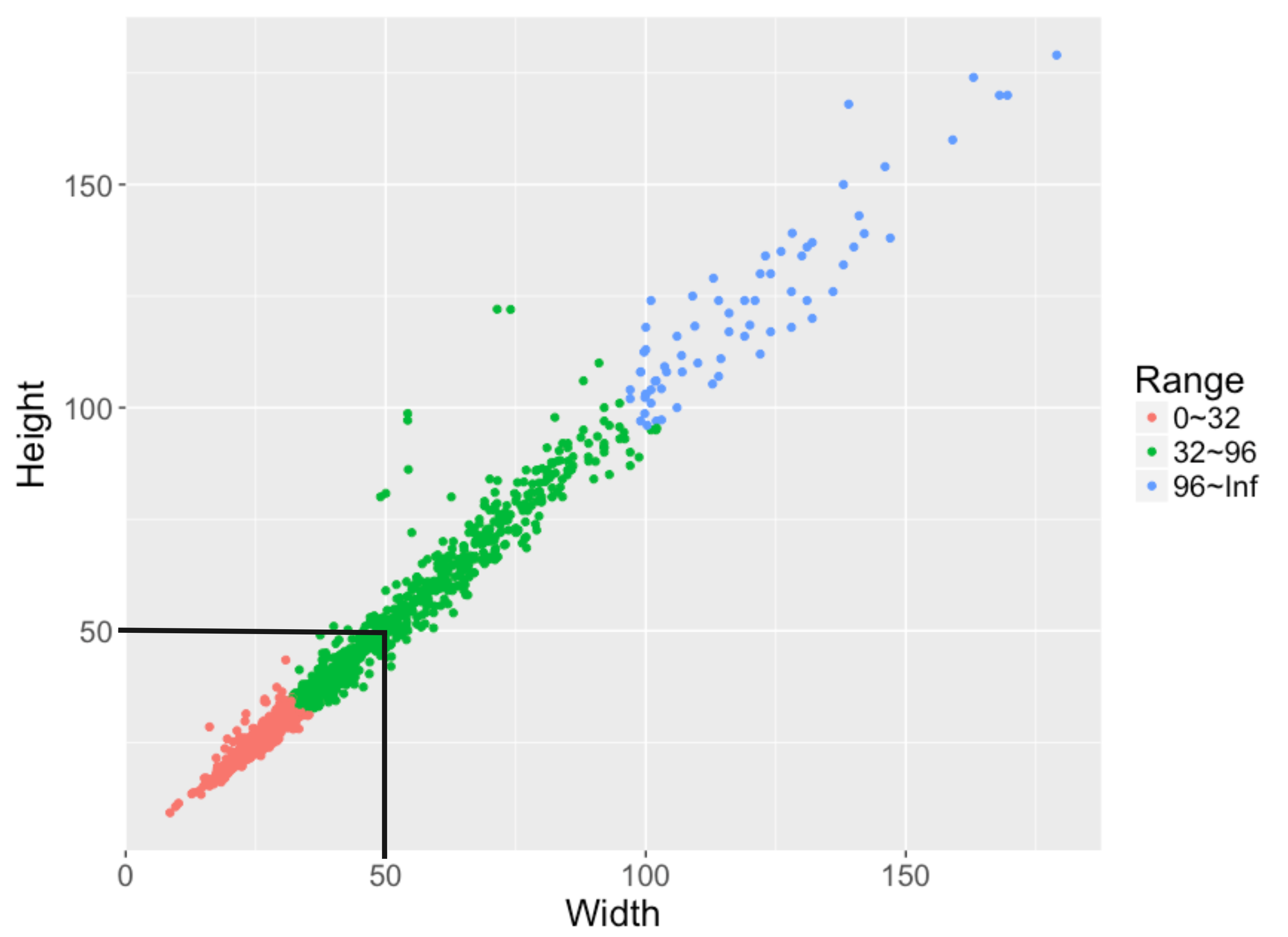}
\caption{Distribution of the traffic signs in STSD testing dataset.}
\label{distribution}
\end{wrapfigure}

We conduct a deep analysis on the STSD dataset and evaluation metrics. Figure \ref{distribution} reports the data distribution of the testing dataset. According to this distribution, most of the traffic signs in the STSD testing dataset are smaller than $50 \times 50$ pixels, only $30.8\%$ traffic signs satisfy the requirements of the evaluation benchmark in ~\cite{STSD}. In other words, the standard STSD benchmark ignores small-scale traffic signs which are widespread in the real world. Detecting small objects is alway a challenge, not only in traffic sign recognition \cite{zhu2016traffic} but also in generic object detection tasks \cite{COCO}. Motivated by the mainstream evaluation metrics in MS COCO \cite{COCO}, we redesign a new evaluation method on the STSD dataset, which takes the wide-range traffic signs into account. In this evaluation method, we incorporate all size of traffic signs into the evaluation and split them into three different scales according to their areas. As shown in Figure \ref{distribution}, $area < 32^2$, $32^2 < area < 96^2$, and $area > 96^2$ denote small, medium and large scale, respectively. In this way, it is possible to evaluate TSR methods based on not only their overall performances but also their multi-scale performances. Moreover, we adopt Average Precision (AP) , the area under the precision-recall curve, to measure class-wise performances. We utilize Average Recall (AR) metric which is an averaged recall over different IoU thresholds ($0.5,0.55,\sim 0.95$) to evaluate region proposal methods.

\textbf{Analysis on DMS-net.} To verify the scale-awareness of our DMS-net, we compare it against three well-known state-of-the-art methods for generic object proposals, including Selective Search \cite{selective_search}, EdgeBoxes \cite{edge-boxes} and RPN \cite{fasterrcnn}. 
Table \ref{proposal_scale} reports the AR values at different proposals and different scales. We can see that our DMS-net is significantly better than the others on all metrics. Our DMS-net also exceeds non-learning methods (Selective Search and EdgeBoxes) by a large margin. Compared with RPN method, we gain $4.6\%$ improvement on both the small-scale and large-scale traffic signs. This proves that our method is effective in detecting multi-scale traffic signs. We also conduct an ablation study by removing the first or second branch but keeping the same number of scales at 10 to verify the effectiveness of the proposed dual multi-scale structure. We observe that if we remove the first branch, the performance on detecting small traffic signs decrease dramatically. The DMS-net without the second branch also has inferior ARs on small and large traffic signs, even though it has a minor increase ($0.2\%$) relative to our entire DMS-net on AR at medium ones. Moreover, We observe that our DMS-net limitedly benefits from the increase in the number of proposals. This means that our method can achieve satisfying results even with few proposals, thereby saving computing time in the later classification task.

\textbf{Analysis on fusion-net.} In order to evaluate our fusion-net, we compare it with the classifier in our baseline method \cite{fcn-edgebox}. For fair comparison, we use the same region proposal method, our DMS-net, to generate proposals for the two classifiers. As we can see from Table \ref{classification}, our fusion-net performs better in most categories of traffic signs and achieves $1.87\%$ mAP higher than the baseline. Moreover, we compare our entire system (DMS-net + fusion-net) with Faster R-CNN. The results in Table \ref{classification} show that our scale-aware TSR outperforms Faster R-CNN by a large margin, $11.67\%$ mAP. We further replace their classifier with our fusion-net, it gains around $10\%$ mAP improvement. This also proves the effectiveness of our fusion-net. For the multi-scale evaluation, our scale-aware TSR obtains high AP scores at different scales. Especially, it achieves $98.1\%$ AP at medium-scale traffic signs.


\begin{table}
\begin{center}
\tiny

\begin{tabular}{lp{0.25cm}p{0.25cm}p{0.25cm}p{0.25cm}p{0.25cm}p{0.25cm}p{0.25cm}p{0.25cm}p{0.25cm}p{0.25cm}p{0.25cm}p{0.25cm}}

    \toprule
    \multirow{1}{*}{Methods}&\multicolumn{1}{c}{ AR@50 }&\multicolumn{1}{c}{ AR@100 }&\multicolumn{1}{c}{ AR@300 }&\multicolumn{1}{c}{ AR@Small }&\multicolumn{1}{c}{ AR@Medium }&\multicolumn{1}{c}{ AR@Large }\cr
    \midrule
    \tiny {Selective Search\cite{selective_search}}
    &4.8	&10.7	&23.4	&1.1	&12.3	&33.4	\cr
    \tiny {EdgeBoxes \cite{edge-boxes}}
	&36.9	&41.0	&47.4	&22.0	&48.3	&67.9	\cr
    \tiny {RPN \cite{fasterrcnn}}
	&66.9	&67.0	&67.2	&57.6	&71.4	&72.1	\cr
    \hline
    \tiny {DMS-net (w/o branch1)}
	&52.5	&52.8	&53.0	&5.7	&74.6	&{\bf77.2}	\cr
    \tiny {DMS-net (w/o branch2)}
	&67.0	& 71.3	&{\bf 72.3}	&61.4	&{\bf 75.8}	&75.3	\cr
    \tiny {DMS-net}
	&{\bf69.8}	&{\bf71.5}	&{\bf72.3}	&{\bf62.2}	&75.6	&76.7	\cr
    \bottomrule
\end{tabular}
\end{center}
\caption{Average Recall (AR) analysis obtained with different proposal numbers and scales. AR for small, medium and large objects are computed for 100 proposals.}
\label{proposal_scale}
\end{table}

\begin{table}[h]
\begin{center}
\tiny
\begin{tabular}{p{1.7cm}|m{0.32cm}<{\centering}m{0.32cm}<{\centering}m{0.32cm}<{\centering}|m{0.32cm}<{\centering}|p{0.32cm}<{\centering}p{0.32cm}<{\centering}p{0.32cm}<{\centering}p{0.32cm}<{\centering}p{0.32cm}<{\centering}p{0.32cm}<{\centering}p{0.32cm}<{\centering}p{0.32cm}<{\centering}p{0.32cm}<{\centering}p{0.32cm}<{\centering}}

Methods & AP (small) & AP (medium) & AP (large) & mAP& {\includegraphics[width=0.38cm]{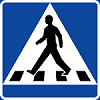}} &{\includegraphics[width=0.38cm]{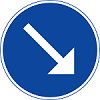}} & {\includegraphics[width=0.38cm]{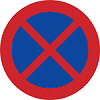}} & {\includegraphics[width=0.38cm]{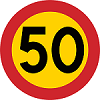}} & {\includegraphics[width=0.38cm]{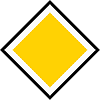}} & {\includegraphics[width=0.38cm]{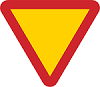}}& {\includegraphics[width=0.38cm]{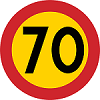}}& {\includegraphics[width=0.38cm]{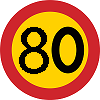}}& {\includegraphics[width=0.38cm]{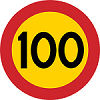}} &{\includegraphics[width=0.38cm]{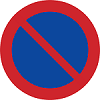}}\\
\hline

DMS-net+ fusion-net & 85.19 &{\bf 98.10} &96.29 & {\bf94.67} &{\bf96.74}	&{\bf98.97}	&{\bf86.67}	&{\bf96.55}	&{\bf95.75}	&90.44	&{\bf98.91}	&{\bf95.45}	&{\bf97.04}	&{\bf90.11} \\

DMS-net+\cite{fcn-edgebox}classifier &82.42 &97.10	&94.15 & 92.80 &96.73	& 98.94	&{\bf 86.67}	& 96.20	& 94.62	& 94.34	&{\bf 98.91}	& 81.65	& 92.90	&87.70 \\

Faster R-CNN\cite{fasterrcnn} & 52.42 &93.96 &{\bf97.72} & 83.00 &86.54 &88.83 &59.32	&81.70	&82.48	&88.24	&91.66	&87.25	&75.64	&88.35 \\

RPN\cite{fasterrcnn}+ fusion-net & {\bf86.63} &97.30 &89.64 & 93.41 &94.65 &97.92 &84.56	&95.98	&90.33	&{\bf94.82}	&97.82 &{\bf95.45} &95.42 &87.13 \\

\end{tabular}
\end{center}
\caption{Average Precision (AP) on all the data and those of different scales and categories. }
\label{classification}
\end{table}









\section{Conclusion}
The paper presented a traffic sign recognition system with performance beyond state-of-the-art methods. The system adopted an FCN with a dual multi-scale CNN architecture to detect traffic signs of different scales, and a concise CNN structure to fuse multi-scale features for classification. Besides, multiple existing effective strategies and modules, e.g. OHEM and Inception, were introduced into the system. The whole system and its modules were evaluated thoroughly on STSD. The system was experimentally proved to be more accurate and faster than state-of-the-art methods. Moreover, the system is more lightweight than the others, thereby more suitable for automobiles with on-board hardware. Future work will be focused on strengthening the system for more challenging tasks, e.g. recognizing traffic signs from images captured in fog weather.

\bibliography{egbib}
\end{document}